# Sub-Pixel Affine Registration of Space Debris Images via the Radon Point Spread Function

SHENSHEN LUAN,[1,2,†] MIAOMIAO TIAN,[2,†] SHUAI JIANG,[2] YAN YANG,[1,*] SHUGUO XIE[1] AND ZEZHOU SUN[2,*]

[1]*School of Electronic and Information Engineering, Beihang University, Beijing 100191, China*
[2]*China Academy of Space Technology, Beijing, 100091, China*
[†]*These authors contributed equally to this work.*
*yanzi@buaa.edu.cn, *sunzezhou1970@126.com

**Inter-frame affine misalignment caused by platform jitter and attitude adjustments poses a fundamental challenge for multi-frame analysis of point targets in optical surveillance. Conventional registration methods rely on spatial intensity correlations or distinctive image features, both of which are largely absent in low-signal-to-noise-ratio point target imagery. We introduce the Radon Point Spread Function (RPSF) to characterize point targets in the Radon-transformed domain, and derive a closed-form framework that jointly estimates inter-frame translation and rotation from as few as four scalar RPSF samples per frame pair. The method requires no iterative optimization, feature extraction or interpolation, which is suitable for resource-constrained onboard processing. Simulation results confirm sub-pixel translation accuracy and a mean rotation error of 0.2556° at 1° Radon angular resolution. Validation on five real space debris datasets including both ground-based and in-orbit observations yields a mean calibration error below 0.5 pixels, substantially exceeding the precision required for reliable multi-frame processing.**

**Introduction.** Optical image sequences are widely employed for space surveillance tasks including debris monitoring and cataloging [1], [2]. Targets of interest typically appear as faint, sub-pixel point sources without discernible textures or shapes, easily indistinguishable from background noise [3]. Multi-frame processing, whether for detection, tracking, or trajectory estimation, depends critically on accurate inter-frame registration. However, subtle platform jitter, mechanical vibrations, and attitude control maneuvers inevitably introduce affine transformations between consecutive frames, predominantly translation and small-angle rotation [4]. Even sub-pixel misalignment can generate spurious residuals in frame-differencing operations, substantially elevating false alarm rates and degrading downstream analysis [5]. Reliable, high-precision inter-frame calibration is therefore a prerequisite for any multi-frame point target processing pipeline. This requirement is especially acute for space-borne platforms, where computational resources are tightly constrained and registration must execute deterministically within fixed time budgets.

Classical intensity-based registration methods estimate inter-frame displacement by maximizing spatial similarity metrics. Phase correlation exploits the Fourier shift property for translation recovery and can be extended to rotation and scale via log-polar mapping [6]. The Enhanced Correlation Coefficient (ECC) algorithm refines parametric motion models through gradient-based optimization of a robust similarity criterion [7]. These approaches perform well on richly textured scenes but degrade sharply when applied to point targets embedded in noise, where spatial correlation is dominated by background fluctuations rather than the target signal. Feature-based methods (e.g., SIFT [8], ORB [9]) are similarly unsuitable, as point targets provide no stable keypoints to match. In astronomy, chi-squared shift estimation has been applied to point-source image registration [10], yet remains iterative in nature and is validated primarily on stellar fields with higher SNR than typical debris imagery. Lu et al. [11] and Nacereddine et al. [12] exploited the affine covariance of the Radon transform to recover transformation parameters, targeting LiDAR-based place recognition and image registration respectively; however, both methods operate on the full Radon sinogram of structured scenes and estimate parameters without addressing the computational constraints relevant to onboard deployment. A closed-form, non-iterative registration method that exploits the sinusoidal signature of individual point targets in the Radon domain and is designed for resource-constrained environments has, to our knowledge, not been reported.

In this work, we introduce the Radon Point Spread Function (RPSF) as a principled framework for point target registration. Because a point target in the image domain maps to a sinusoidal curve in the Radon domain, the affine transformation parameters, translation magnitude and direction and rotation angle, are algebraically encoded in the amplitude, phase, and horizontal shift of this sinusoid. Unlike prior Radon-domain affine estimators that operate on the full sinogram of structured scenes [11], [12], we recover inter-frame parameters from the RPSF curves of only two reference points, requiring four scalar RPSF samples from the reference frame and a one-dimensional angular scan of the target frame's difference curve. Crucially, once the per-frame Radon transform is computed, all subsequent parameter recovery is purely algebraic and non-iterative, yielding deterministic execution time. The method is validated through controlled simulations covering sub-pixel to multi-pixel translations and a full 0° to 180° rotation range, and further tested on five real space debris datasets acquired from ground-based and in-orbit platforms. Experimental results demonstrate mean calibration errors below 0.5 pixels.

**Method and principle.** The Radon Transform is a rotational integration operation in the transform domain, where a pixel in the image domain corresponds to a sine curve in the Radon domain. The position of pixels in the image domain has a one-to-one correspondence with the amplitude and initial phase of the sine curve, as illustrated in Fig. 1.

Define the system function of the Radon transform as (1), which we call the Radon Point Spread Function.

$$\rho(\theta)_{[x,y]} = x\cdot\cos\theta + y\cdot\sin\theta \quad (1)$$

Where, [x, y] are the pixel coordinates in the image domain, θ is the rotation angle in the Radon domain, and $\rho(\theta)_{[x,y]}$ represents the amplitude of the corresponding sine function. For the Radon transform of point targets, $\rho(\theta)$ can be determined by taking the ρ coordinate value corresponding to the maximum value in each column of the Radon domain.

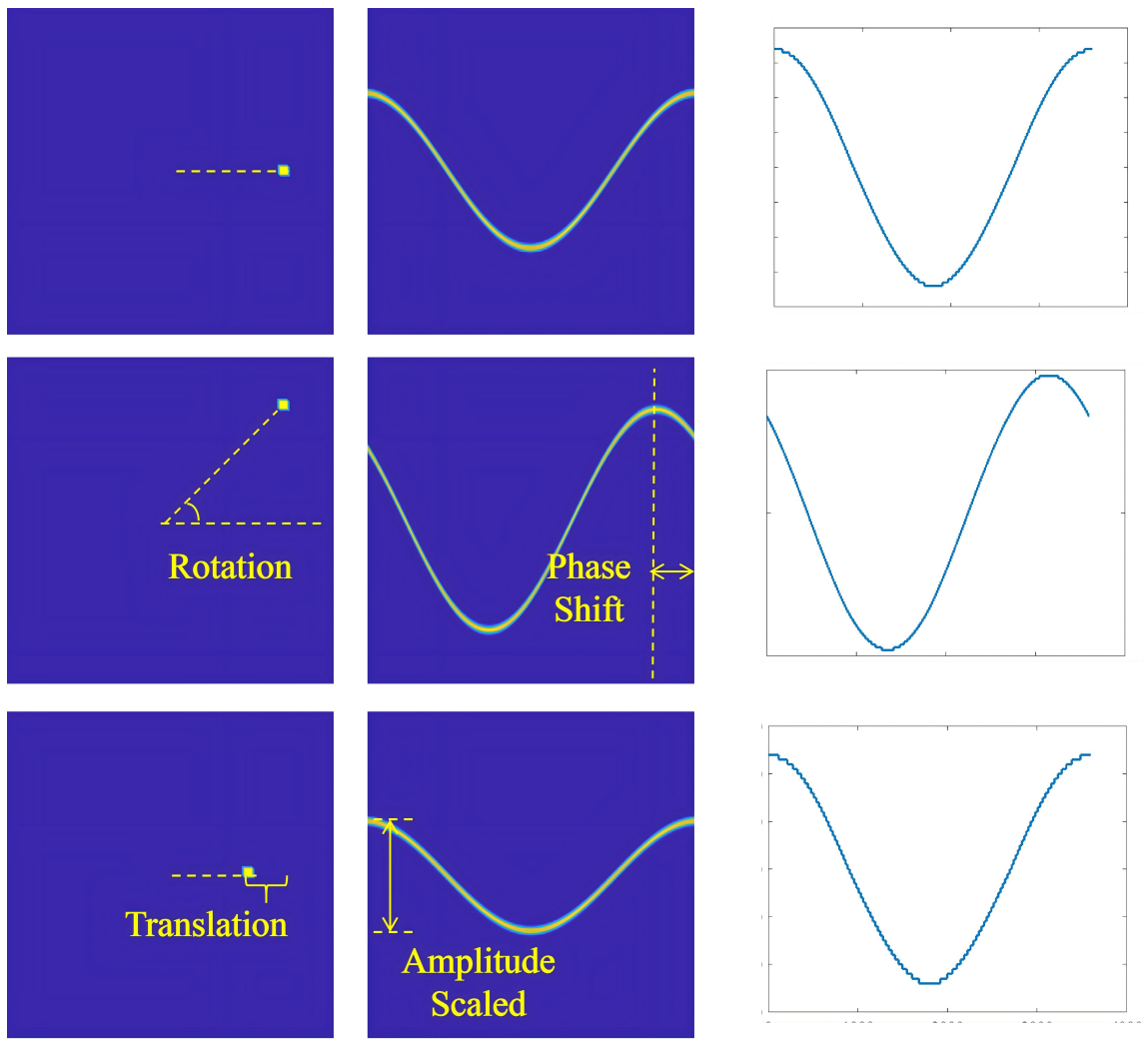

Fig. 1. The correlation between the position of pixels and the amplitude and initial phase of the sine curve .

Therefore, the translation, rotation, and scaling transformations of the point target Radon transform in Supplement 1 can be represented on the Radon point spread function as shown in (2), (3), and (4). By combining them with (1), the translation parameter [l, ψ ], rotation parameter β, and scaling parameter A can be solved.

$$\rho(\theta)_{\mathrm{Trans}[(x,y)]} = x\cdot\cos\theta + y\cdot\sin\theta + l\cdot\cos(\theta-\psi) \quad (2)$$

$$\rho(\theta)_{\mathrm{Rotate}[(x,y)]} = x\cdot\cos(\theta+\beta) + y\cdot\sin(\theta+\beta) \quad (3)$$

$$\rho(\theta)_{\mathrm{Scale}[(x,y)]} = A\cdot x\cdot\cos\theta + A\cdot y\cdot\sin\theta \quad (4)$$

For applications such as space debris detection, there is usually only translation and rotation transformation between consecutive frame images. Therefore, the RPSF of joint translation and rotation transformations can be derived as following.

$$\rho(\theta)_{\mathrm{R\&T}[(x,y)]} = \rho(\theta+\beta)_{\mathrm{T}[(x,y)]} = x\cdot\cos(\theta+\beta) + y\cdot\sin(\theta+\beta) + l\cdot\cos(\theta+\beta-\psi) \quad (5)$$

Let $\tau = \theta+\beta$, then the above equation can be further written as the following one.

$$\rho(\tau)_{\mathrm{T}[(x,y)]} = x\cdot\cos(\tau) + y\cdot\sin(\tau) + l\cdot\cos(\tau-\psi) \quad (6)$$

According to (1), a certain pixel $(x_0, y_0)$ of the original image can be represented by the RPSF as:

$$\rho(\theta)_{[(x_0,y_0)]} = x_0\cdot\cos\theta + y_0\cdot\sin\theta = L_0\cdot\cos(\theta-\alpha_0) \quad (7)$$

Wherein, $L_0 = \sqrt{x_0^2+y_0^2}$, $\alpha_0 = \arctan\frac{y_0}{x_0}$. Then, (6) can be written as:

$$\rho(\tau)_{\mathrm{T}[(x,y)]} = L_0\cdot\cos(\tau-\alpha_0) + l\cdot\cos(\tau-\psi) \quad (8)$$

As described before, the affine transformation composed of translation and rotation can be solved for the respective parameters by combining (7) and (8).

Without loss of generality, we define the RPSF of point targets before and after affine transformation as $\rho_1(\theta)_{[(x_0,y_0)]}$ and $\rho_2(\theta+\beta_0)_{\mathrm{T}[(x_0,y_0)]}$. According to (7) and (8), it can be inferred that:

$$\rho_1(\theta)_{[(x_0,y_0)]} = L\cdot\cos(\theta-\alpha) \quad (9)$$

$$P_2(\theta+\beta_0)_{\mathrm{T}[(x_0,y_0)]} = L\cdot\cos(\theta+\beta_0-\alpha) + l\cdot\cos(\theta+\beta_0-\psi) \quad (10)$$

Where $l = \sqrt{t_{x_0}^2+t_{y_0}^2}$, $\psi = \arctan\frac{t_{y_0}}{t_{x_0}}$ and $t_{x_0}$, $t_{y_0}$ and $\beta_0$ are the translation distance in both directions and rotation angle between the two frames, respectively. The purpose of interframe image matching and object detection for debris is to obtain the values of $t_{x_0}$, $t_{y_0}$ and $\beta_0$ by jointly solving functions $\rho_1$ and $\rho_2$. Here we develop the estimating methods of rotation angle and translation distance for the inter-frame image calibration described in the Supplement 2. The main solution is now given as following. Evaluating four scalar RPSF values, the vector of the affine transformation can ge obtained in closed form:

$$\psi = \beta_0 - \arctan\frac{-\left(\rho_2\left(\frac{\pi}{2}+\beta_0\right)_{\mathrm{T}[(x_1,y_1)]} - \rho_1\left(\frac{\pi}{2}+\beta_0\right)\right)}{\rho_2(\beta_0)_{\mathrm{T}[(x_1,y_1)]} - \rho_1(\beta_0)} \quad (11)$$

$$l = \sqrt{[\rho_2(\beta_0)_{\mathrm{T}[(x_1,y_1)]} - \rho_1(\beta_0)]^2 + [\rho_2\left(\frac{\pi}{2}+\beta_0\right)_{\mathrm{T}[(x_1,y_1)]} - \rho_1\left(\frac{\pi}{2}+\beta_0\right)]^2} \quad (12)$$

The Cartesian translation vector is $(t_x, t_y) = (l\cdot\cos\psi, l\cdot\sin\psi)$. A second reference point provides an independent estimate that can be averaged for improved robustness. All step-by-step derivations are provided in Supplement 2.

**Simulation results.** To validate the translation estimation, a 512 × 512 synthetic image containing a single 2D Gaussian point target ($\sigma$=1.5 px) with additive white Gaussian noise (SNR = 10 dB) was generated, emulating a challenging low-SNR observation. Translation vectors $(t_x, t_y)$ spanning from -1.6 to 1.6 pixels with a 0.1 pixel step were applied, covering both integer and sub–pixel displacements.The Radon transform was computed with an angular step size $\Delta\theta = 1°$. For each translation vector, the parameters $(\hat{t}_x, \hat{t}_y)$ were estimated using the proposed method and the Euclidean error $\Delta t = \sqrt{(\hat{t}_x - t_x)^2 + (\hat{t}_y - t_y)^2}$ was recorded.

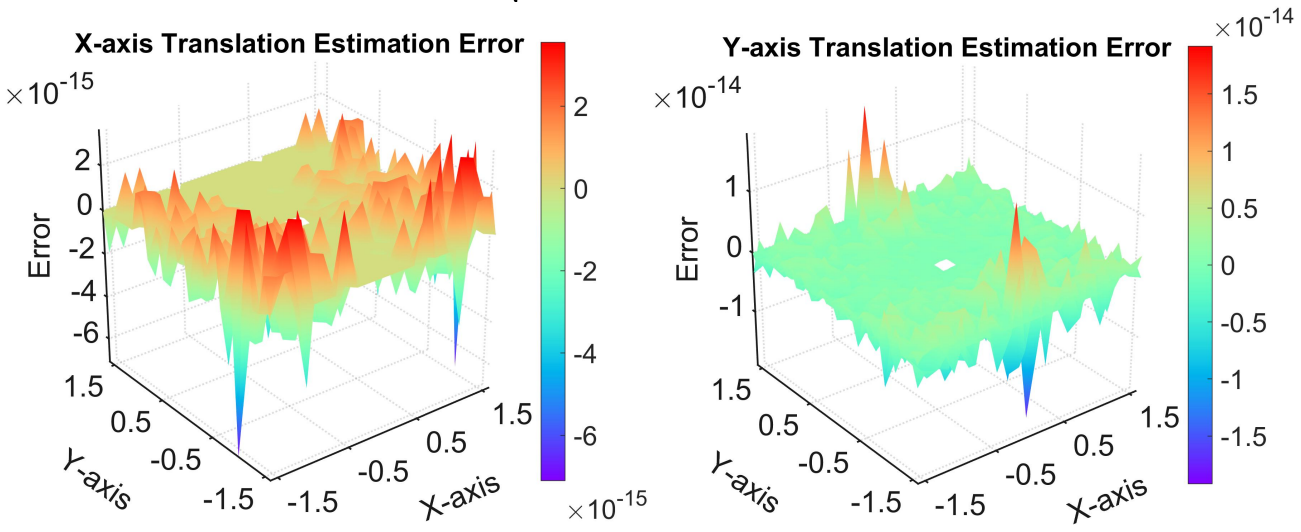

Fig. 2. Translation estimation error on X-axis and Y-axis.The heat map depicts the Euclidean error (in pixels) for translation vectors (t_x, t_y) ranging from -1.6 to 1.6 pixels. The data point at the origin (0,0) is intentionally omitted, as the frame difference contains no usable signal for estimation under zero displacement.

The missing data point at the origin (0,0) in Fig. 2 corresponds to the degenerate case of zero inter-frame displacement, where the frame difference contains no coherent target signal and the RPSF signature is absent. Across all remaining displacement vectors, the mean Euclidean error was nearly 0 pixels, confirming sub-pixel accuracy at SNR = 10 dB. The error exhibits no systematic dependence on the direction or magnitude of the displacement, indicating the isotropy of the RPSF-based estimator.

For rotation estimation, the synthetic image was rotated from 0° to 180° at 1° step. The Radon transform was computed at three angular resolutions ( $\Delta\theta = 1°, 0.1°, 0.01°$ ). Two reference points were positioned at distances exceeding 200 pixels from the rotation center to provide a long effective lever arm for RPSF-based measurement. The statistical summary for all three angular resolutions is presented in Table 1. As $\Delta\theta$ decreases from 1° to 0.01°, the mean rotation error drops from 0.2556° to 0.0042°, consistent with the expectation that finer Radon angular sampling yields more precise estimates. We note that the 0.0042° accuracy represents an idealized upper bound achievable under noise-free simulation conditions.

**Table 1. Statistical results of rotation angle estimation errors**

| Radon Transform Accuracy(°) | | 0.01 | 0.1 | 1 |
|---|---|---|---|---|
| Error (°) | Average | 0.0042 | 0.0591 | 0.2556 |
| | Maximum | 0.05 | 0.15 | 0.5 |
| | Minimum | 0 | 0 | 0 |

The rationale for selecting distant reference points is quantitatively validated . We fixed the rotation angle at 1 ° and systematically varied the distance of the reference points from the rotation center during estimation. The results are shown in Fig. 3, where the horizontal axis represents the distance between the reference point and the rotation center, and the vertical axis represents the estimation error. It is evident that the further the reference point is from the rotation center, the smaller the error becomes. Additionally, a smaller step size for the Radon transform correlates with reduced error. For a Radon transform with a step size of 1°, the reference point should be at least 200 pixels away from the rotation center to minimize the impact of errors.

**Experimental results.** We evaluate the calibration accuracy of the proposed method on five real datasets of space debris optical imagery.

The datasets comprise observations from two sources: (1) ground-based images from the National Astronomical Observatory of China (Datasets 1–3), and (2) in-orbit images acquired by the optical payload aboard the Tianzhou-III spacecraft (Datasets 4–5). Dataset specifications are summarized in Table 2. Owing to the short inter-frame interval and the Earth-oriented three-axis stabilized attitude, the inter-frame transformations involve only translation, with negligible rotational components. The ground-truth translation vectors, listed in Table 2, range from sub-pixel to tens of pixels across different datasets.

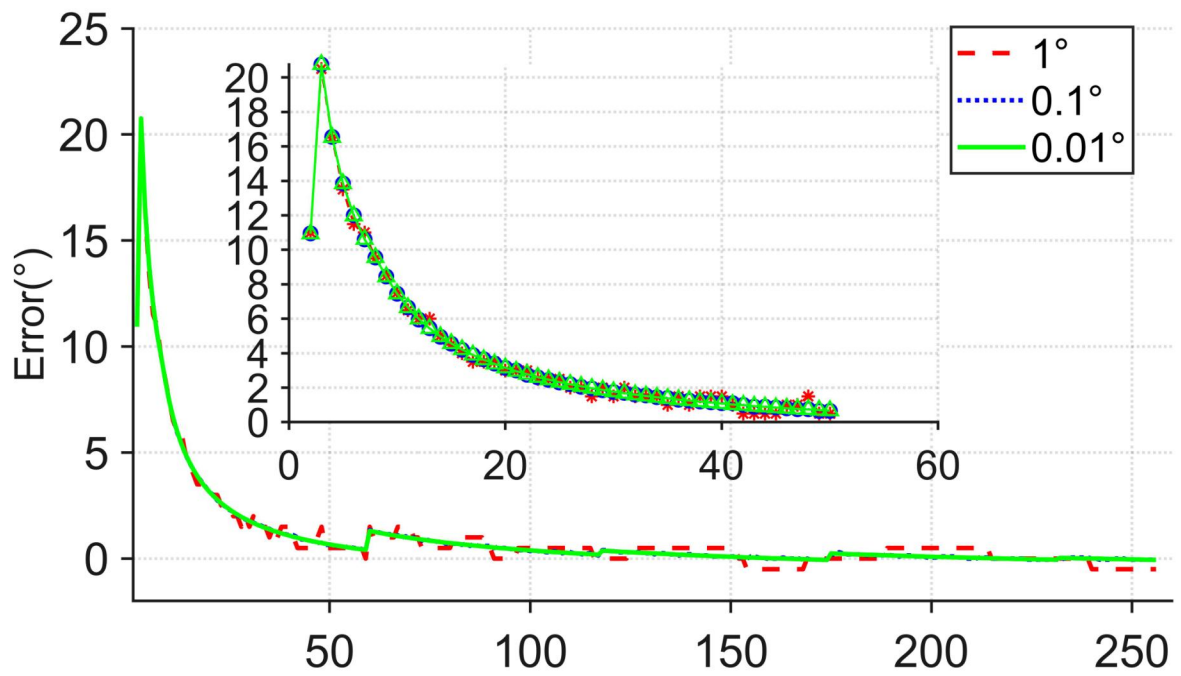

Fig. 3. Rotation estimation error versus the distance of the reference point from the rotation center. The error decreases monotonically as the distance increases. For a Radon step size of 1° , a distance of at least 200 pixels is required to minimize the error impact.

Fig. 4 compares the mean translation error of the proposed RPSF method and three baselines, for the x- and y-axes separately. The ground-truth displacements range from sub-pixel to tens of pixels. The RPSF method attains a mean error below 0.75 pixels on every dataset and axis, with an overall average below 0.5 pixels and a maximum of about 3 pixels, which occurs in low-SNR frames. Phase correlation returns near-zero estimates on the large-displacement datasets and mean errors of 13–65 pixels on the others. ECC returns 1–3 pixels on D3–D5 but 31–55 pixels on D1–D2, as its capture range is limited to about 32 pixels. Chi2 shift returns 0.09 – 1.86 pixels but mis-estimates a ~1-pixel shift as ±1024 pixels on two D5 pairs, owing to the periodic-boundary assumption of FFT correlation. RPSF maintains sub-pixel accuracy on all datasets. Fig. 5 illustrates the estimated inter-frame translation vectors for representative frame pairs across the five datasets. The vector directions and magnitudes vary across datasets, reflecting differences in observational geometry and platform motion.

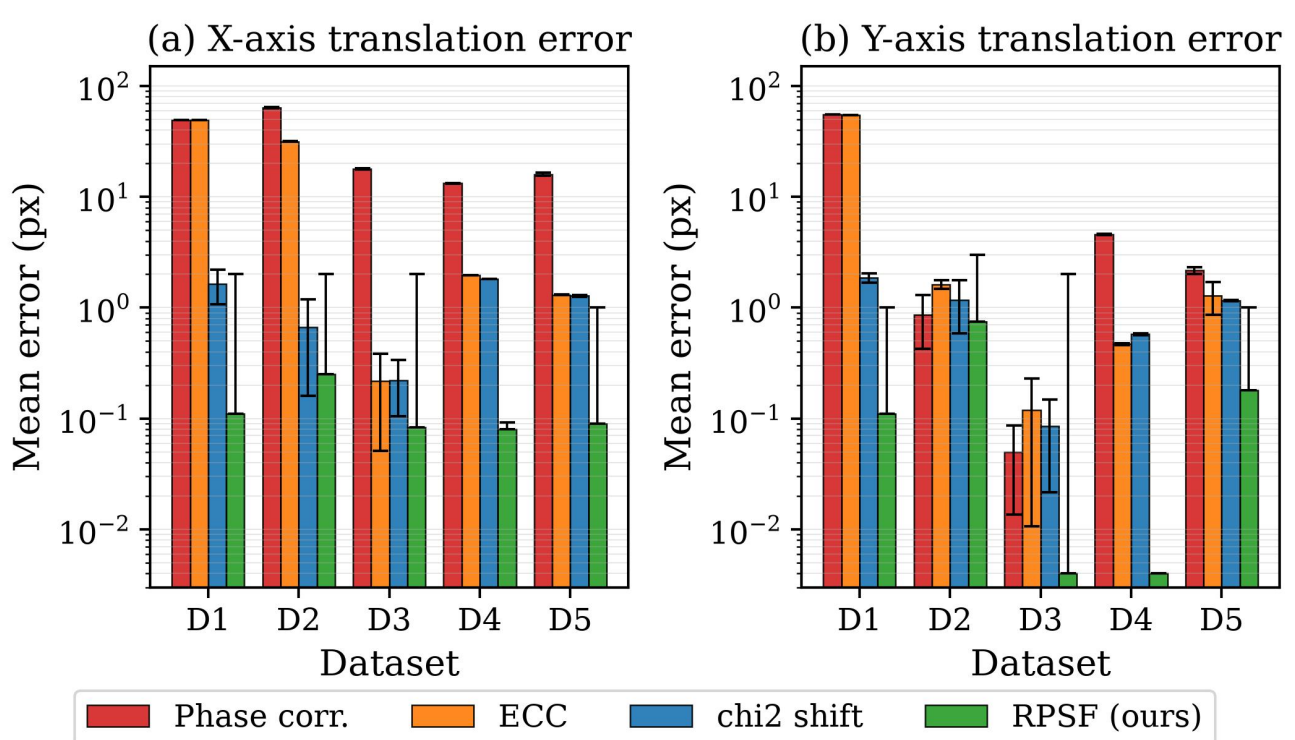

Fig. 4. Comparison of mean translation error (pixels) among phase correlation, ECC, chi2 shift, and the proposed RPSF method on five real datasets.

To contextualize the calibration accuracy, we compared the proposed RPSF-based method against three established registration approaches, including phase correlation [6], ECC [7], and chi-squared shift estimation [10] on the same five datasets. For each dataset, the ground-truth translation vector (Table 2) served as the reference. The mean Euclidean error , averaged over all frame pairs within each dataset, is reported for each method.

Table 2. Comparison of translation estimation errors (mean Euclidean error in pixels) across five datasets.

| Dataset (capacity) (GT shift) | Axis | Phase corr.[6] | ECC [7] | chi2 shift [10] | RPSF (ours) |
|---|---|---|---|---|---|
| D1 (18) (-49, 55) | x | 49.00 / 49.00 | 49.04 / 49.11 | 1.64 / 3.33 | **0.11 / 2.00** |
| | y | 55.00 / 55.00 | 54.86 / 54.97 | 1.86 / 2.12 | **0.11 / 1.00** |
| D2 (21) (-32, 1) | x | 63.55 / 65.53 | 31.59 / 31.79 | 0.67 / **1.97** | **0.25** / 2.00 |
| | y | 0.86 / 1.87 | 1.63 / 1.89 | 1.17 / 3.05 | **0.75 / 3.00** |
| D3 (13) (9, 0) | x | 17.79 / 18.38 | 0.22 / 0.70 | 0.22 / **0.50** | **0.08** / 2.00 |
| | y | 0.05 / 0.15 | 0.12 / 0.36 | 0.09 / **0.20** | **0.00** / 2.00 |
| D4 (10) (-6, -2) | x | 13.21 / 13.26 | 1.97 / 1.97 | 1.82 / 1.82 | **0.08 / 0.09** |
| | y | 4.60 / 4.74 | 0.47 / 0.48 | 0.58 / 0.59 | **0.00 / 0.00** |
| D5 (45) (-7, -1) | x | 16.03 / 16.19 | 1.31 / 1.33 | 1.27 / 1.29 | **0.09 / 1.00** |
| | y | 2.17† / — | 1.28 / 3.21 | 1.16† / — | **0.18 / 1.00** |

As summarized in Table 2, phase correlation fails on every dataset, returning either a zero shift on the large-displacement dataset D1 (49-55 px mean error) or spurious correlation peaks (15-65 px), because low-SNR point targets provide no reliable intensity-correlation surface. ECC is accurate for small displacements (D3-D5) but collapses to a near-zero solution once the displacement exceeds roughly 32 px (D1, D2). The chi2-shift method attains the best baseline accuracy (0.09-1.86 px) yet suffers a catastrophic cyclic ambiguity on two frame pairs of D5, where a 1 px shift is mis-estimated as ±1024 px owing to the periodic-boundary assumption of FFT correlation. In contrast, the proposed RPSF method maintains a mean error below 0.75 px and a maximum error below 3 px across all five datasets in both axes, confirming its robustness to low SNR, large displacement, and cyclic ambiguity.

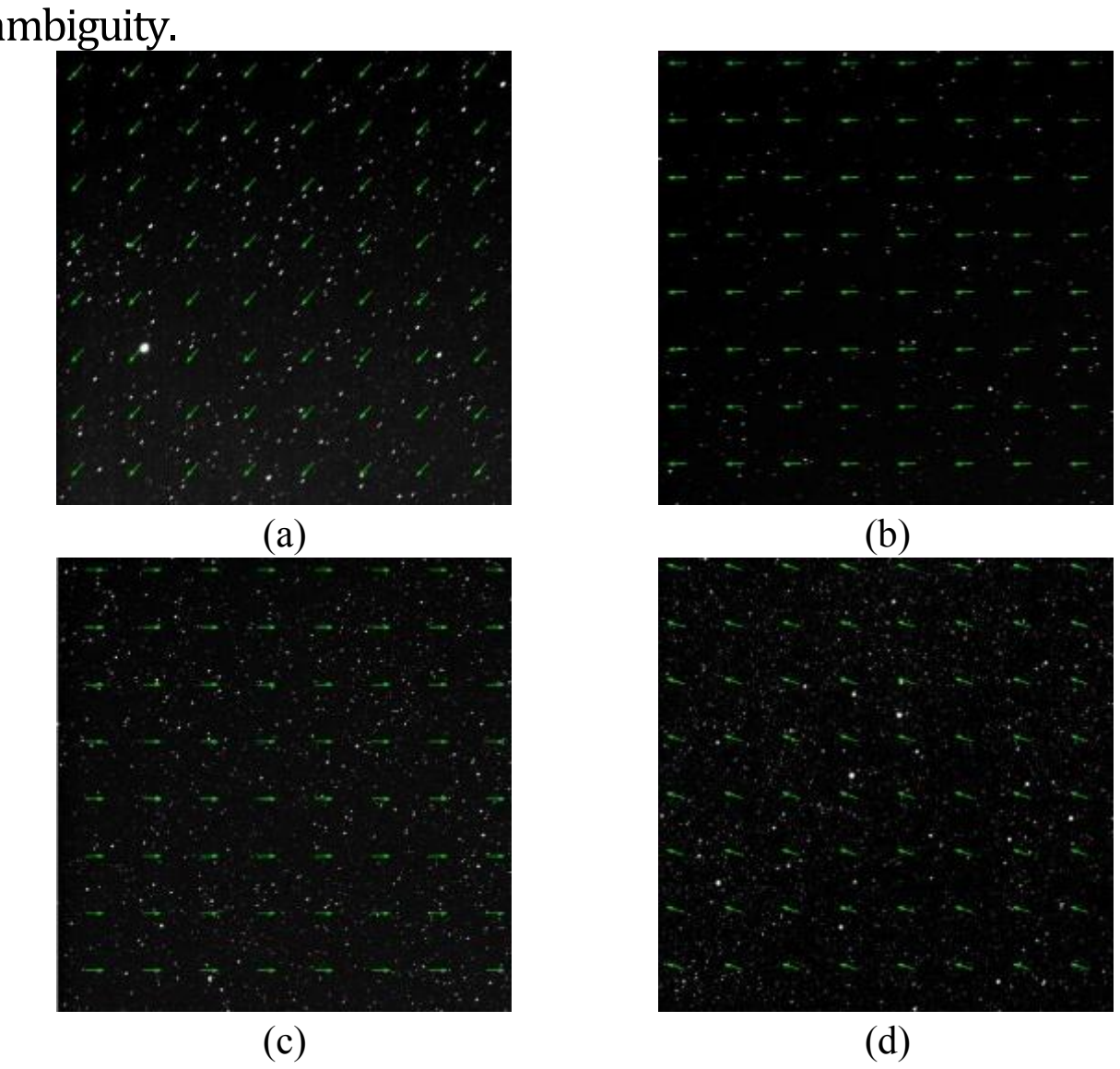

(a) (b) (c) (d)

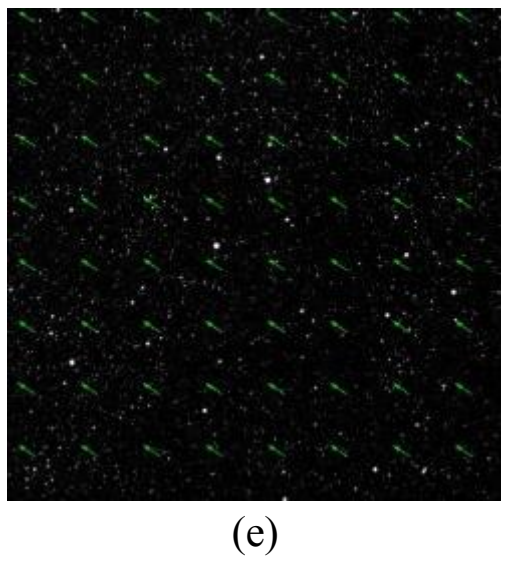

(e)

Fig. 5. The translation vector of the dataset 1-5.

**Conclusion.** We have presented a closed-form framework for sub-pixel affine registration of point target images via the Radon Point Spread Function. Once the per-frame Radon transform is computed, the affine parameters are recovered algebraically in deterministic, non-iterative time, making the method suitable for resource-constrained onboard deployment. Simulations confirm sub-pixel translation accuracy and a mean rotation error of 0.2556°, while validation on five real space debris datasets yields a mean calibration error below 0.5 pixels, establishing the RPSF framework as an accurate and lightweight registration solution.

**Back Matter**

**Funding.** National Natural Science Foundation of China (62401034).

**Disclosures**. The authors declare no conflicts of interest.

**Data Availability Statement (DAS).** Data underlying the results presented in this paper are not publicly available at this time but may be obtained from the authors upon reasonable request.

**Supplemental Document.** See Supplement 1 and 2 for supporting content.

## References(with titles, for review only)